\documentclass[10pt,twocolumn,letterpaper]{article}

\usepackage{iccv}

\makeatletter
\@namedef{ver@everyshi.sty}{}
\makeatother

\usepackage[dvipsnames,svgnames,x11names]{xcolor}
\usepackage{tikz}
\usetikzlibrary{arrows.meta,shapes,calc,matrix,fit,positioning,backgrounds,decorations.markings,fadings}
\usepackage{pgfplots}
\usepackage{pgfplotstable}
\pgfplotsset{compat=1.9}
\usepackage{xstring}

\usepgfplotslibrary{external}

\IfBeginWith*{\jobname}{fig/extern/}{\finalcopy}{}

\tikzstyle{every picture}+=[
	remember picture,
	every text node part/.style={align=center},
	every matrix/.append style={ampersand replacement=\&},
]
\tikzstyle{tight} = [inner sep=0pt,outer sep=0pt]
\tikzstyle{node}  = [draw,circle,tight,minimum size=12pt,anchor=center]
\tikzstyle{op}    = [draw,circle,tight]
\tikzstyle{dot}   = [fill,draw,circle,inner sep=1pt,outer sep=0]
\tikzstyle{pt}    = [fill,draw,circle,inner sep=1.5pt,outer sep=.2pt]
\tikzstyle{box}   = [draw,thick,rectangle,inner sep=3pt]
\tikzstyle{high}  = [black!60]
\tikzstyle{group} = [high,box,opacity=.5]
\tikzstyle{dim1}  = [fill opacity=.3,text opacity=1]
\tikzstyle{dim2}  = [fill opacity=.5,text opacity=1]
\tikzstyle{dim3}  = [fill opacity=.7,text opacity=1]
\tikzstyle{rectc} = [tight,transform shape]
\tikzstyle{rect}  = [rectc,anchor=south west]

\tikzset{every mark/.append style={solid}}
\pgfplotsset{
	grid=both, width=\columnwidth, try min ticks=5,
	every axis/.append style={font=\small},
	every axis plot/.append style={thick,mark=none,mark size=1.8,tension=0.18},
	legend cell align=left, legend style={fill opacity=0.8},
	xticklabel={\pgfmathprintnumber[assume math mode=true]{\tick}},
	yticklabel={\pgfmathprintnumber[assume math mode=true]{\tick}},
	nodes near coords math/.style={
		nodes near coords={\pgfmathprintnumber[assume math mode=true]{\pgfplotspointmeta}},
	},
}

\pgfplotsset{
	dash/.style={mark=o,dashed,opacity=0.6},
	dott/.style={mark=o,dotted,opacity=0.6},
	nolim/.style={enlargelimits=false},
	plain/.style={every axis plot/.append style={},nolim,grid=none},
}

\pgfdeclarelayer{bg4}
\pgfdeclarelayer{bg3}
\pgfdeclarelayer{bg2}
\pgfdeclarelayer{bg1}
\pgfdeclarelayer{fg1}
\pgfdeclarelayer{fg2}
\pgfdeclarelayer{fg3}
\pgfdeclarelayer{fg4}
\pgfsetlayers{bg4,bg3,bg2,bg1,main,fg1,fg2,fg3,fg4}

\pgfkeys{/tikz/.cd, aspect/.store in=\aspect, aspect=1}
\pgfkeys{/tikz/.cd, depth/.store in=\depth, depth=.5}
\pgfkeys{/tikz/.cd, stepx/.store in=\step, stepx=1}

\tikzstyle{geom} = [line join=bevel,aspect=1,depth=.5,z={(\depth*\aspect,\depth)}]
\tikzstyle{wire} = [geom,draw,thick]

\def\cx[#1,#2,#3]{#1}
\def\cy[#1,#2,#3]{#2}
\def\cz[#1,#2,#3]{#3}
\def\ex[#1,#2,#3]{#1,0,0}
\def\ey[#1,#2,#3]{0,#2,0}
\def\ez[#1,#2,#3]{0,0,#3}

\newcommand{\zrect}[3][]{%
\path[geom,#1] #2 rectangle +(\cx[#3],\cy[#3]);
}
\newcommand{\yrect}[3][]{%
\path[geom,#1,shift={#2},xslant=\aspect]
	(0,0) rectangle +(\cx[#3],\depth*\cz[#3]);
}
\newcommand{\xrect}[3][]{%
\path[geom,#1,shift={#2},yslant=1/\aspect]
	(0,0) rectangle +(\aspect*\depth*\cz[#3],\cy[#3]);
}

\newcommand{\para}[4][]{%
\zrect[#1]{(#3)}{#4}                 
\yrect[#1]{($(#3)+(\ey[#4])$)}{#4}   
\xrect[#1]{($(#3)+(\ex[#4])$)}{#4}   
\path[geom]
	(#3) coordinate(#2-southwest)
	($(#3)+(#4)$) coordinate(#2-northeast)
	($(#3)+(\ey[#4])$) coordinate(#2-northwest)
	($(#3)+(#4)-(\ey[#4])$) coordinate(#2-southeast)
	($(#3)+.5*(\ex[#4])$) coordinate(#2-south)
	($(#3)+(#4)-.5*(\ex[#4])$) coordinate(#2-north)
	($(#3)+.5*(\ex[#4])+.5*(#4)$) coordinate(#2-center)
	(#2-southwest |- #2-center) coordinate(#2-west)
	(#2-center -| #2-northeast) coordinate(#2-east)
	;
}

\makeatletter
\renewcommand\paragraph{\@startsection{paragraph}{4}{\z@}{1ex}{-1em}{\normalfont\normalsize\bfseries}}
\makeatother

\usepackage{times}
\usepackage{epsfig}
\usepackage{graphicx}
\usepackage{amsmath}
\usepackage{amssymb}

\usepackage{times}
\usepackage{helvet}
\usepackage{courier}
\usepackage{url}
\usepackage{graphicx}
\usepackage{booktabs}
\usepackage{subcaption}
\usepackage{amsmath}
\usepackage{amsfonts}
\usepackage{mathtools}
\usepackage{multirow}
\usepackage{enumitem}

\usepackage{times}
\usepackage{epsfig}
\usepackage{graphicx}
\usepackage{amsmath}
\usepackage{amssymb}
\usepackage{tabularx}

\usepackage{verbatim}
\usepackage{multirow}
\usepackage{array}

\usepackage{amsthm}
\usepackage{xspace}
\usepackage[svgnames]{xcolor}

\usepackage{pifont}

\usepackage[pagebackref=true,breaklinks=true,letterpaper=true,colorlinks,bookmarks=false]{hyperref}

\iccvfinalcopy

\begin{document}

\title{All the attention you need:\\Global-local, spatial-channel attention for image retrieval}

\author{Chull Hwan Song\\
{\small Odd Concepts}
\and
Hye Joo Han\\
{\small Odd Concepts}
\and
Yannis Avrithis\\
{\small Inria, Univ Rennes, CNRS, IRISA}
}

\maketitle


\newcommand{\head}[1]{{\smallskip\noindent\textbf{#1}}}
\newcommand{\alert}[1]{{\color{red}{#1}}}
\newcommand{\sm}{\scriptsize}
\newcommand{\eq}[1]{(\ref{eq:#1})}

\newcommand{\Th}[1]{\textsc{#1}}
\newcommand{\mr}[2]{\multirow{#1}{*}{#2}}
\newcommand{\mc}[2]{\multicolumn{#1}{c}{#2}}
\newcommand{\mca}[3]{\multicolumn{#1}{#2}{#3}}
\newcommand{\tb}[1]{\textbf{#1}}
\newcommand{\ch}{\checkmark}

\newcommand{\red}[1]{{\color{red}{#1}}}
\newcommand{\blue}[1]{{\color{blue}{#1}}}
\newcommand{\green}[1]{\color{green}{#1}}
\newcommand{\gray}[1]{{\color{gray}{#1}}}

\newcommand{\citeme}[1]{\red{[XX]}}
\newcommand{\refme}[1]{\red{(XX)}}

\newcommand{\fig}[2][1]{\includegraphics[width=#1\linewidth]{fig/#2}}
\newcommand{\figh}[2][1]{\includegraphics[height=#1\linewidth]{fig/#2}}


\newcommand{\tran}{^\top}
\newcommand{\mtran}{^{-\top}}
\newcommand{\zcol}{\mathbf{0}}
\newcommand{\zrow}{\zcol\tran}

\newcommand{\ind}{\mathbbm{1}}
\newcommand{\expect}{\mathbb{E}}
\newcommand{\nat}{\mathbb{N}}
\newcommand{\zahl}{\mathbb{Z}}
\newcommand{\real}{\mathbb{R}}
\newcommand{\proj}{\mathbb{P}}
\newcommand{\prob}{\mathbf{Pr}}
\newcommand{\normal}{\mathcal{N}}

\newcommand{\mif}{\textrm{if}\ }
\newcommand{\other}{\textrm{otherwise}}
\newcommand{\minimize}{\textrm{minimize}\ }
\newcommand{\maximize}{\textrm{maximize}\ }
\newcommand{\st}{\textrm{subject\ to}\ }

\newcommand{\id}{\operatorname{id}}
\newcommand{\const}{\operatorname{const}}
\newcommand{\sgn}{\operatorname{sgn}}
\newcommand{\var}{\operatorname{Var}}
\newcommand{\mean}{\operatorname{mean}}
\newcommand{\trace}{\operatorname{tr}}
\newcommand{\diag}{\operatorname{diag}}
\newcommand{\vect}{\operatorname{vec}}
\newcommand{\cov}{\operatorname{cov}}
\newcommand{\sign}{\operatorname{sign}}
\newcommand{\prj}{\operatorname{proj}}

\newcommand{\sigmoid}{\operatorname{sigmoid}}
\newcommand{\softmax}{\operatorname{softmax}}
\newcommand{\clip}{\operatorname{clip}}

\newcommand{\defn}{\mathrel{:=}}
\newcommand{\peq}{\mathrel{+\!=}}
\newcommand{\meq}{\mathrel{-\!=}}

\newcommand{\floor}[1]{\left\lfloor{#1}\right\rfloor}
\newcommand{\ceil}[1]{\left\lceil{#1}\right\rceil}
\newcommand{\inner}[1]{\left\langle{#1}\right\rangle}
\newcommand{\norm}[1]{\left\|{#1}\right\|}
\newcommand{\abs}[1]{\left|{#1}\right|}
\newcommand{\frob}[1]{\norm{#1}_F}
\newcommand{\card}[1]{\left|{#1}\right|\xspace}
\newcommand{\diff}{\mathrm{d}}
\newcommand{\der}[3][]{\frac{d^{#1}#2}{d#3^{#1}}}
\newcommand{\pder}[3][]{\frac{\partial^{#1}{#2}}{\partial{#3^{#1}}}}
\newcommand{\ipder}[3][]{\partial^{#1}{#2}/\partial{#3^{#1}}}
\newcommand{\dder}[3]{\frac{\partial^2{#1}}{\partial{#2}\partial{#3}}}

\newcommand{\wb}[1]{\overline{#1}}
\newcommand{\wt}[1]{\widetilde{#1}}

\def\xssp{\hspace{1pt}}
\def\ssp{\hspace{3pt}}
\def\msp{\hspace{5pt}}
\def\lsp{\hspace{12pt}}

\newcommand{\cA}{\mathcal{A}}
\newcommand{\cB}{\mathcal{B}}
\newcommand{\cC}{\mathcal{C}}
\newcommand{\cD}{\mathcal{D}}
\newcommand{\cE}{\mathcal{E}}
\newcommand{\cF}{\mathcal{F}}
\newcommand{\cG}{\mathcal{G}}
\newcommand{\cH}{\mathcal{H}}
\newcommand{\cI}{\mathcal{I}}
\newcommand{\cJ}{\mathcal{J}}
\newcommand{\cK}{\mathcal{K}}
\newcommand{\cL}{\mathcal{L}}
\newcommand{\cM}{\mathcal{M}}
\newcommand{\cN}{\mathcal{N}}
\newcommand{\cO}{\mathcal{O}}
\newcommand{\cP}{\mathcal{P}}
\newcommand{\cQ}{\mathcal{Q}}
\newcommand{\cR}{\mathcal{R}}
\newcommand{\cS}{\mathcal{S}}
\newcommand{\cT}{\mathcal{T}}
\newcommand{\cU}{\mathcal{U}}
\newcommand{\cV}{\mathcal{V}}
\newcommand{\cW}{\mathcal{W}}
\newcommand{\cX}{\mathcal{X}}
\newcommand{\cY}{\mathcal{Y}}
\newcommand{\cZ}{\mathcal{Z}}

\newcommand{\vA}{\mathbf{A}}
\newcommand{\vB}{\mathbf{B}}
\newcommand{\vC}{\mathbf{C}}
\newcommand{\vD}{\mathbf{D}}
\newcommand{\vE}{\mathbf{E}}
\newcommand{\vF}{\mathbf{F}}
\newcommand{\vG}{\mathbf{G}}
\newcommand{\vH}{\mathbf{H}}
\newcommand{\vI}{\mathbf{I}}
\newcommand{\vJ}{\mathbf{J}}
\newcommand{\vK}{\mathbf{K}}
\newcommand{\vL}{\mathbf{L}}
\newcommand{\vM}{\mathbf{M}}
\newcommand{\vN}{\mathbf{N}}
\newcommand{\vO}{\mathbf{O}}
\newcommand{\vP}{\mathbf{P}}
\newcommand{\vQ}{\mathbf{Q}}
\newcommand{\vR}{\mathbf{R}}
\newcommand{\vS}{\mathbf{S}}
\newcommand{\vT}{\mathbf{T}}
\newcommand{\vU}{\mathbf{U}}
\newcommand{\vV}{\mathbf{V}}
\newcommand{\vW}{\mathbf{W}}
\newcommand{\vX}{\mathbf{X}}
\newcommand{\vY}{\mathbf{Y}}
\newcommand{\vZ}{\mathbf{Z}}

\newcommand{\va}{\mathbf{a}}
\newcommand{\vb}{\mathbf{b}}
\newcommand{\vc}{\mathbf{c}}
\newcommand{\vd}{\mathbf{d}}
\newcommand{\ve}{\mathbf{e}}
\newcommand{\vf}{\mathbf{f}}
\newcommand{\vg}{\mathbf{g}}
\newcommand{\vh}{\mathbf{h}}
\newcommand{\vi}{\mathbf{i}}
\newcommand{\vj}{\mathbf{j}}
\newcommand{\vk}{\mathbf{k}}
\newcommand{\vl}{\mathbf{l}}
\newcommand{\vm}{\mathbf{m}}
\newcommand{\vn}{\mathbf{n}}
\newcommand{\vo}{\mathbf{o}}
\newcommand{\vp}{\mathbf{p}}
\newcommand{\vq}{\mathbf{q}}
\newcommand{\vr}{\mathbf{r}}
\newcommand{\Vs}{\mathbf{s}}
\newcommand{\vt}{\mathbf{t}}
\newcommand{\vu}{\mathbf{u}}
\newcommand{\vv}{\mathbf{v}}
\newcommand{\vw}{\mathbf{w}}
\newcommand{\vx}{\mathbf{x}}
\newcommand{\vy}{\mathbf{y}}
\newcommand{\vz}{\mathbf{z}}

\newcommand{\vone}{\mathbf{1}}
\newcommand{\vzero}{\mathbf{0}}

\newcommand{\valpha}{{\boldsymbol{\alpha}}}
\newcommand{\vbeta}{{\boldsymbol{\beta}}}
\newcommand{\vgamma}{{\boldsymbol{\gamma}}}
\newcommand{\vdelta}{{\boldsymbol{\delta}}}
\newcommand{\vepsilon}{{\boldsymbol{\epsilon}}}
\newcommand{\vzeta}{{\boldsymbol{\zeta}}}
\newcommand{\veta}{{\boldsymbol{\eta}}}
\newcommand{\vtheta}{{\boldsymbol{\theta}}}
\newcommand{\viota}{{\boldsymbol{\iota}}}
\newcommand{\vkappa}{{\boldsymbol{\kappa}}}
\newcommand{\vlambda}{{\boldsymbol{\lambda}}}
\newcommand{\vmu}{{\boldsymbol{\mu}}}
\newcommand{\vnu}{{\boldsymbol{\nu}}}
\newcommand{\vxi}{{\boldsymbol{\xi}}}
\newcommand{\vomikron}{{\boldsymbol{\omikron}}}
\newcommand{\vpi}{{\boldsymbol{\pi}}}
\newcommand{\vrho}{{\boldsymbol{\rho}}}
\newcommand{\vsigma}{{\boldsymbol{\sigma}}}
\newcommand{\vtau}{{\boldsymbol{\tau}}}
\newcommand{\vupsilon}{{\boldsymbol{\upsilon}}}
\newcommand{\vphi}{{\boldsymbol{\phi}}}
\newcommand{\vchi}{{\boldsymbol{\chi}}}
\newcommand{\vpsi}{{\boldsymbol{\psi}}}
\newcommand{\vomega}{{\boldsymbol{\omega}}}

\newcommand{\rLambda}{\mathrm{\Lambda}}
\newcommand{\rSigma}{\mathrm{\Sigma}}

\newcommand{\vLambda}{\bm{\rLambda}}
\newcommand{\vSigma}{\bm{\rSigma}}

\makeatletter
\newcommand*\bdot{\mathpalette\bdot@{.7}}
\newcommand*\bdot@[2]{\mathbin{\vcenter{\hbox{\scalebox{#2}{$\m@th#1\bullet$}}}}}
\makeatother

\makeatletter
\DeclareRobustCommand\onedot{\futurelet\@let@token\@onedot}
\def\@onedot{\ifx\@let@token.\else.\null\fi\xspace}

\def\eg{\emph{e.g}\onedot} \def\Eg{\emph{E.g}\onedot}
\def\ie{\emph{i.e}\onedot} \def\Ie{\emph{I.e}\onedot}
\def\cf{\emph{cf}\onedot} \def\Cf{\emph{Cf}\onedot}
\def\etc{\emph{etc}\onedot} \def\vs{\emph{vs}\onedot}
\def\wrt{w.r.t\onedot} \def\dof{d.o.f\onedot} \def\aka{a.k.a\onedot}
\def\etal{\emph{et al}\onedot}
\makeatother


\def\vlad{VLAD\xspace}
\def\smk{SMK$^{\star}$\xspace}
\def\asmk{ASMK$^{\star}$\xspace}
\def\sp{SP\xspace}
\def\qe{QE\xspace}
\def\hqe{HQE\xspace}
\def\dfs{DFS\xspace}
\def\off{O}
\def\hesaff{HesAff\xspace}
\def\rsift{rSIFT\xspace}
\def\delf{DELF\xspace}


\def\oxf5k{Ox5k\xspace}
\def\paris6k{Par6k\xspace}

\def\roxf{$\mathcal{R}$Oxford}
\def\rox{$\mathcal{R}$Oxf}
\def\ro{$\mathcal{R}$O}
\def\rpar{$\mathcal{R}$Paris}
\def\rpa{$\mathcal{R}$Par}
\def\rp{$\mathcal{R}$Pe}
\def\r1m{$\mathcal{R}$1M}
\def\rs{$\mathcal{R}$100k}


\definecolor{greenn}{rgb}{0.30,0.69,0.31}
\definecolor{redd}{rgb}{0.3,0.2,0.7}
\definecolor{red}{rgb}{0.8,0.1,0.1}

\newcommand{\ok}[1]{\color{redd}{#1}}
\newcommand{\okg}[1]{\color{greenn}{#1}}

\begin{abstract}
We address representation learning for large-scale instance-level image retrieval. Apart from backbone, training pipelines and loss functions, popular approaches have focused on different spatial pooling and attention mechanisms, which are at the core of learning a powerful global image representation. There are different forms of attention according to the interaction of elements of the feature tensor (local and global) and the dimensions where it is applied (spatial and channel). Unfortunately, each study addresses only one or two forms of attention and applies it to different problems like classification, detection or retrieval.

We present \emph{global-local attention module} (GLAM), which is attached at the end of a backbone network and incorporates all four forms of attention: local and global, spatial and channel. We obtain a new feature tensor and, by spatial pooling, we learn a powerful embedding for image retrieval.
Focusing on global descriptors, we provide empirical evidence of the interaction of all forms of attention and improve the state of the art on standard benchmarks.
\end{abstract}

\section{Introduction}
\label{sec:intro}

\begin{figure*}
\centering
\tikzfading[
	name=fade out,
	inner color=transparent!90,
	outer color=transparent!10,
]

\begin{tikzpicture}[
	scale=.3,
	font={\footnotesize},
	node distance=.5,
	ovr/.style={fill=white,fill opacity=.9},
	ten/.style={draw,ovr},
	ops/.style={op,ovr},
	rec/.style args={(#1/#2)}{draw,rectc,minimum width=#1cm,minimum height=#2cm,preaction={ovr}},
	att/.style={fill,path fading=fade out},
	atth/.style={att,fading transform={yscale=10}},
	dim/.style={text opacity=.5,inner sep=2pt,below=2pt of #1},
	sym/.style={above=3pt of #1},
	symt/.style={above=8pt of #1-northwest},
	key/.style={red,left,near end},
	back/.style={draw=#1!60,fill=#1!30,fill opacity=.5,inner sep=4},
	xback/.style={back=#1,inner xsep=8},
	yback/.style={back=#1,inner ysep=8},
]
\matrix[
	tight,
	row sep={38,between origins},column sep=12,
	cells={scale=.3,},
	nodes={node distance=.5},
] {
	\&[4]\&
	\node[rec=(4/1),atth] (lc) {};
	\node[sym=lc] {$\vA_c^l$};
	\node[dim=lc] {$c \times 1 \times 1$};
	\&
	\node[ops] (lc1) {$\times$};
	\&
	\node[ops] (lc2) {$+$};
	\&
	\para[ten]{lcf}{0,-2.5,-2}{1,5,4};
	\node[symt=lcf] {$\vF_c^l$};
	\&\&
	\node[rec=(4/4),att] (ls) {};
	\node[sym=ls] {$\vA_s^l$};
	\node[dim=ls] {$1 \times h \times w$};
	\node[ops,right=of ls] (ls1) {$\times$};
	\&
	\node[ops] (ls2) {$+$};
	\&
	\para[ten]{lsf}{0,-2.5,-2}{1,5,4};
	\node[symt=lsf] {$\vF^l$};
	\&[5]
	\node[ops] (l1) {$\times$};
	\\
	\para[ten]{if}{0,-2.5,-2}{1,5,4};
	\node[ovr,dim=if-south] {$c \times h \times w$};
	\node[symt=if] {$\vF$};
	\&
	\node[dot](s) at(0,0){};
	\&\&\&\&\&
	\node[dot] (c){};
	\&\&\&\&
	\node[ops] (f1) {$\times$};
	\&
	\node[ops] (f) {$+$};
	\&
	\para[ten]{of}{0,-2.5,-2}{1,5,4};
	\node[dim=of-south] {$c \times h \times w$};
	\node[symt=of] {$\vF^{gl}$};
	\\
	\&\&
	\node[rec=(4/4),att] (gc) {};
	\node[sym=gc] {$\vA_c^g$};
	\node[dim=gc] {$c \times c$};
	\&
	\node[ops] (gc1) {$\times$};
	\&\&
	\para[ten]{gcf}{0,-2.5,-2}{1,5,4};
	\node[symt=gcf] {$\vF_c^g$};
	\&\&
	\node[rec=(5/5),att] (gs) {};
	\node[sym=gs] {$\vA_s^g$};
	\node[dim=gs] {$hw \times hw$};
	\node[ops,right=of gs] (gs1) {$\times$};
	\&
	\node[ops] (gs2) {$+$};
	\&
	\para[ten]{gsf}{0,-2.5,-2}{1,5,4};
	\node[symt=gsf] {$\vF^g$};
	\&
	\node[ops] (g1) {$\times$};
	\\
};

\node[above=of l1] (wl) {$w_l$};
\node[above=of f1] (w)  {$w$};
\node[above=of g1] (wg) {$w_g$};

\draw (if-east)--(s);
\draw[->] (s) |- (lc);
\draw[->] (s) |- (gc);
\draw[->] (l1) -| (f);
\draw[->] (g1) -| (f);

\draw[->]
	(lc) edge (lc1)
	(lc1) edge (lc2)
	(lc2) edge (lcf-west)
	(ls) edge (ls1)
	(ls1) edge (ls2)
	(ls2) edge (lsf-west)
	(lsf-east)--(l1)
	;

\draw[->]
	(s)--(f1)
	(f1) edge (f)
	(f) edge (of-west)
	;

\draw[->] (c) |- (ls);
\draw[->] (c) |- (gs);

\draw[->]
	(gc) edge (gc1)
	(gc1) edge (gcf-west)
	(gs) edge (gs1)
	(gs1) edge (gs2)
	(gs2) edge (gsf-west)
	(gsf-east)--(g1)
	;

\draw (lcf-north) -- +(0,.6) coordinate (lcf-n);
\draw[->] (lcf-n) -| (ls1);
\draw[->] (lcf-n) -| (ls2);

\draw (gcf-south) -- +(0,-.6) coordinate (gcf-s);
\draw[->] (gcf-s) -| (gs1);
\draw[->] (gcf-s) -| (gs2);

\path
	(s |- lcf-n) coordinate(s-n)
	(s |- gcf-s) coordinate(s-s)
	;
\draw[->] (s-n) -| (lc1);
\draw[->] (if-north) |- (s-n) -| (lc2);
\begin{pgfonlayer}{bg1}
	\draw[->] (if-south) |- (s-s) -| (gc1);
\end{pgfonlayer}

\draw[->]
	(wl) edge (l1)
	(w) edge (f1)
	(wg) edge (g1)
	;

\begin{pgfonlayer}{bg2}
	\node[yback=blue,fit=(s-n) (s-s) (lc2)] (channel) {};
	\node[yback=red,fit=(lcf-n -| ls1) (gcf-s -| gs1) (c) (gs2)] (spatial) {};
	\node[back=black,fit=(l1) (g1) (f)] (fusion) {};
	\node[xback=yellow,fit=(s-n) (lcf-n) (lcf-south) (lsf-east)] (local) {};
	\node[xback=green,fit=(s-s) (gcf-north) (gcf-s) (gsf-east)] (global) {};
\end{pgfonlayer}

\node[blue,below=1pt of channel]{channel attention};
\node[red,below=1pt of spatial]{spatial attention};
\node[black,below=1pt of fusion]{fusion};
\node[yellow!60!red,above=1pt of local.north east]{local attention};
\node[green!60!black,below=1pt of global.south east]{global attention};

\end{tikzpicture}
\caption{Our \emph{global-local attention module} (GLAM) involves both {\color{blue}channel} and {\color{red}spatial} attention, as well as both {\color{yellow!60!red}local} attention (channels/locations weighted independently, based on contextual information obtained by pooling) and {\color{green!60!black}global} attention (based on pairwise interaction between channels/locations). As a result, four attention maps are used: \emph{local channel} ($\vA_c^l$), \emph{local spatial} ($\vA_s^l$), \emph{global channel} ($\vA_c^g$) and \emph{global spatial} ($\vA_s^g$). The input feature map $\vF$ is weighted into local ($\vF^l$) and global ($\vF^g$) attention feature maps, which are fused with $\vF$ to yield the \emph{global-local attention feature map} $\vF^{gl}$. The diagram is abstract: The four attention modules are shown in more detail in Figures \ref{fig:fig4}, \ref{fig:fig3}, \ref{fig:fig6}, \ref{fig:fig5}.}
\label{fig:glam}
\end{figure*}
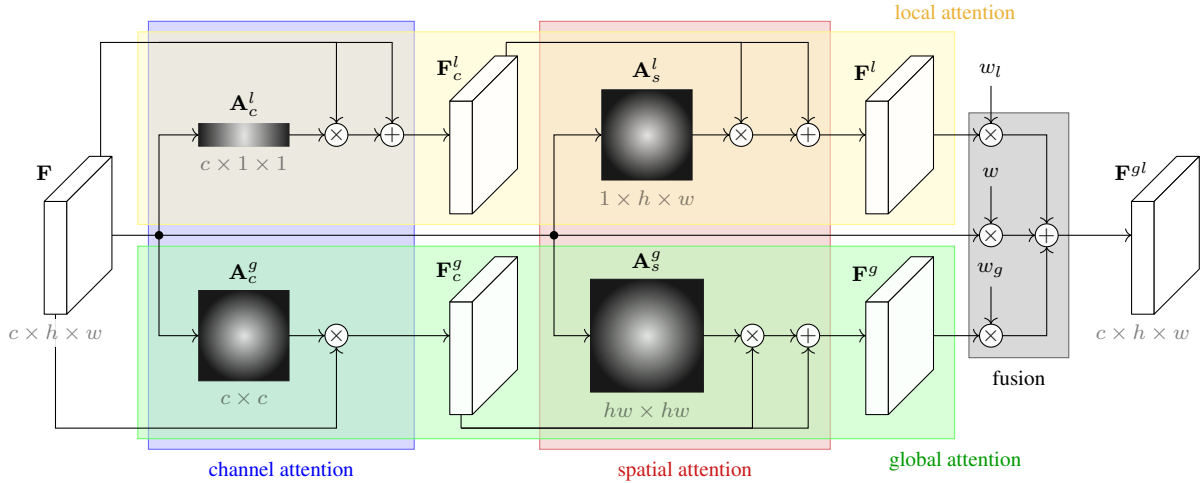

Instance-level image retrieval is at the core of visual representation learning and is connected with many problems of visual recognition and machine learning, for instance \emph{metric learning}~\cite{oh2016deep,KKCK20}, \emph{few-shot learning}~\cite{SnellSZ17} and \emph{unsupervised learning}~\cite{chen2020simple}. Many large-scale open datasets~\cite{Babenko01, Radenovic01, Gordo01, Noh01, Weyand01}, and competitions\footnote{https://www.kaggle.com/c/landmark-retrieval-2020} have accelerated progress in instance-level image retrieval, which has been transformed by deep learning~\cite{Babenko01}.

Many studies on instance-level image retrieval focus on learning features from \emph{convolutional neural networks} (CNN), while others focus on \emph{re-ranking}, for instance by graph-based methods~\cite{Donoser01}. The former can be distinguished according to feature types: \emph{local descriptors}, reminiscent of SIFT~\cite{Lowe01}, where an image is mapped to a few hundred vectors; and \emph{global descriptors}, where an image is mapped to a single vector. In fact, deep learning has brought global descriptors with astounding performance, while allowing efficient search. Our study belongs to this type.

Studies on global descriptors have focused on \emph{spatial pooling}~\cite{Babenko03,Radenovic01}. The need for compact, discriminative representations that are resistant to clutter has naturally given rise to \emph{spatial attention} methods~\cite{Kalantidis01,Ng01}. Different kinds of attention have been studied in many areas of computer vision research. There is also \emph{channel attention}~\cite{Hu01,ChenKLYF18}; \emph{local attention}, applied independently to elements of the representation (feature map)~\cite{woo01,Kim01}; \emph{global attention}, based on interaction between elements~\cite{Wang02,ChenKLYF18}; and combinations thereof. Unfortunately, each study has been limited to one or two kinds of attention only; attention is not always learned; and applications vary.

It is the objective of our work to perform a comprehensive study of all forms of attention above, apply them to instance-level image retrieval and provide a detailed account of their interaction and impact on performance. As shown in \autoref{fig:glam}, we collect contextual information from images with both \emph{local} and \emph{global} attention, giving rise to two parallel network streams. Importantly, each operates on both \emph{spatial locations} and \emph{feature channels}. Local attention is about individual locations and channels; global is about interaction between locations and between channels. The extracted information is separately embedded in local and global attention feature maps, which are combined in a \emph{global-local attention feature map} before pooling.

Our contributions can be summarized as follows:
\begin{enumerate}[itemsep=2pt, parsep=0pt, topsep=0pt]
	 \item We propose a novel network that consists of both global and local attention for image retrieval. This is the first study that employs both mechanisms.
	 \item Each of the global and local attention mechanisms comprises both spatial and channel attention.
	 \item Focusing on global descriptors, we provide empirical evidence of the interaction of all forms of attention and improve the state of the art on standard benchmarks.
\end{enumerate}

\section{Related work}
\label{sec:related}

\paragraph{Instance-level image retrieval}

Studies on instance-level image retrieval can be roughly, but not exclusively, divided into three types: (1) studies on \emph{global descriptors} \cite{Babenko01, Gordo01, Kalantidis01, Weyand01, Babenko03, Radenovic01}; (2) studies on \emph{local descriptors} and geometry-based re-ranking \cite{Noh01, Teichmann01, simeoni2019local, Weyand01}; (3) \emph{re-ranking} by graph-based methods \cite{Donoser01, iscen2017efficient, Yang01}.
The first two types of studies focus on the feature representation, while the last type focuses on re-ranking extracted features.

Studies on global descriptors focus on \emph{spatial pooling} of CNN feature maps into vectors, including MAC~\cite{Razavian2015VisualIR}, SPoC~\cite{Babenko03}, CroW~\cite{Kalantidis01}, R-MAC~\cite{ToliasSJ15, Gordo00, Gordo01}, GeM~\cite{Radenovic01}, and NetVLAD~\cite{Arandjelovic01, Kim01}, as well as \emph{learning the representation}~\cite{Babenko01, Gordo00, Gordo01, Radenovi01, Radenovic01}. Studies before deep learning dominated image retrieval were mostly based on \emph{local descriptors} like SIFT~\cite{Lowe01} and \emph{bag-of-words} representation~\cite{Philbin01} or aggregated descriptors like VLAD~\cite{JPD+11} or ASMK~\cite{TAJ13}. Local descriptors have been revived in deep learning, \eg with DELF~\cite{Noh01}, DELG~\cite{ECCV2020_912} and ASMK extensions~\cite{Teichmann01, tolias2020learning}.

\begin{table}
\centering
\small
\setlength{\tabcolsep}{2.6pt}
\begin{tabular}{lcccccc} \toprule
	 \mr{2}{\Th{Method}}                          & \mc{2}{\Th{Local}}    & \mc{2}{\Th{Global}} & \mr{2}{\Th{Lrn}} & \mr{2}{\Th{Ret}} \\ \cmidrule{2-5}
	                                              & Spatial   & Channel   & Spatial  & Channel  &                  &                  \\ \midrule
	 SENet~\cite{Hu01}                            &           & \ch       &          &          & \ch              &                  \\
	 ECA-Net~\cite{wang01}                        &           & \ch       &          &          & \ch              &                  \\
	 GCNet~\cite{Cao01}                           &           & \ch       &          &          & \ch              &                  \\
	 CBAM~\cite{woo01}                            & \ch       & \ch       &          &          & \ch              &                  \\
	 GE~\cite{HuSASV18}                           & \ch       &           &          &          & \ch              &                  \\
	 NL-Net~\cite{Wang02}                         &           &           & \ch      &          & \ch              &                  \\
	 AA-Net~\cite{Bello_2019_ICCV}                &           &           & \ch      &          & \ch              &                  \\
	 SAN~\cite{zhao2020exploring}                 &           &           & \ch      &          & \ch              &                  \\
	 N$^3$Net~\cite{plotz2018neural}              &           &           & \ch      &          & \ch              &                  \\
	 A$^2$-Net~\cite{ChenKLYF18}                  &           &           &          & \ch      & \ch              &                  \\
	 GSoP~\cite{Gao_2019_CVPR}                    &           &           &          & \ch      & \ch              &                  \\ \midrule
	 OnA~\cite{JimenezAN17}                       & \ch       &           &          &          &                  & \ch              \\
	 AGeM~\cite{gu2018attention}                  & \ch       &           &          &          &                  & \ch              \\
	 CroW~\cite{Kalantidis01}                     & \ch       & \ch       &          &          &                  & \ch              \\
	 CRN~\cite{Kim01}                             & \ch       &           &          &          & \ch              & \ch              \\
	 DELF~\cite{Noh01}                            & \ch       &           &          &          & \ch              & \ch              \\
	 DELG~\cite{ECCV2020_912}                     & \ch       &           &          &          & \ch              & \ch              \\
	 Tolias \etal~\cite{tolias2020learning}       & \ch       &           &          &          & \ch              & \ch              \\
	 SOLAR~\cite{Ng01}                            &           &           & \ch      &          & \ch              & \ch              \\ \midrule
	 \tb{Ours}                                    & \ch       & \ch       & \ch      & \ch      & \ch              & \ch              \\ \bottomrule
\end{tabular}
\caption{Related work on attention. LRN: learned; RET: applied to instance-level image retrieval.}
\label{tab:rel}
\end{table}

We focus on learning a global descriptor in this work, because it is the most efficient in terms of storage and search. However, our generic attention mechanism produces a feature tensor and could be applicable to local descriptors as well, if global pooling were replaced by local feature detection. Re-ranking methods are complementary to the representation and we do not consider them in this work.

\tikzstyle{lay} = [draw,minimum width=50pt,inner sep=4pt]
\tikzstyle{inp} = [lay,fill=black!15]
\tikzstyle{gap} = [lay,fill=brown!25]
\tikzstyle{down} = [lay,fill=yellow!15,trapezium,trapezium angle=110]
\tikzstyle{up} = [lay,fill=yellow!15,trapezium,trapezium angle=70]
\tikzstyle{fix} = [lay,fill=blue!10]
\tikzstyle{cat} = [lay,fill=green!15]
\tikzstyle{fun} = [lay,ellipse,fill=orange!25]
\tikzstyle{outp} = [lay,fill=red!15]
\tikzstyle{dim} = [black!50]
\tikzstyle{key} = [red]


\begin{figure}
\centering
\begin{tikzpicture}[
	font={\footnotesize},
]
\matrix[
	row sep=15pt,column sep=10pt,cells={scale=1,},
]{
	\node[inp](i){feature map}; \\
	\node[gap](g){GAP}; \\
	\node[fix](c5){conv1d($k$)}; \\
	\node[fun](s){$\sigmoid$}; \\
	\node[outp](o){attention map}; \\
};
\draw[->] 
	(i) edge node[dim,midway,right]{$c \times h \times w$} (g)
	(g) edge node[dim,midway,right]{$c \times 1 \times 1$} (c5) 
	(c5) edge (s)
	(s) edge node[dim,midway,right]{$c \times 1 \times 1$} (o)
	;
\node[left=1pt of i] {$\vF$};
\node[left=1pt of o] {$\vA_c^l$};
\end{tikzpicture}
\caption{Local channel attention.}
\label{fig:fig4}
\end{figure}
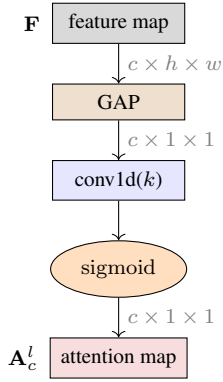

\paragraph{Attention}

Attention mechanisms have been first proposed in \emph{image classification} studies focusing on \emph{channel attention}~\cite{Hu01, wang01, Cao01}, \emph{spatial attention}~\cite{HuSASV18} or both, like CBAM~\cite{woo01}. In \emph{image retrieval}, CroW \cite{Kalantidis01} also  employs both spatial and channel attention and can be seen as a precursor of CBAM, but, like other studies of spatial attention on retrieval~\cite{simeoni2019graph, JimenezAN17, gu2018attention}, it is not learned. CRN~\cite{Kim01} applies spatial attention for feature reweighting and is learned. Learned spatial attention mechanisms are common for local descriptors~\cite{Noh01, ECCV2020_912, tolias2020learning}.

We call the above methods \emph{local attention}, in the sense that elements of the feature tensor (channels / spatial locations), are weighted independently, based on contextual information obtained by pooling or learned. By constrast, by \emph{global attention} we refer to mechanisms that model interaction between elements of the feature tensor, for example between channels or between locations.

In \emph{image classification}, \emph{non-local neural network} (NLNet)~\cite{Wang02} is maybe the first global attention mechanism, followed by similar studies~\cite{Bello_2019_ICCV,zhao2020exploring,plotz2018neural}. It is global \emph{spatial attention}, allowing interaction between any pair of spatial locations. Similarly, there are studies of global \emph{channel attention}, allowing interaction between channels~\cite{ChenKLYF18, Gao_2019_CVPR}. Global attention has focused mostly on image recognition and has been applied to either spatial or channel attention so far, not both. In \emph{image retrieval}, SOLAR~\cite{Ng01} is a direct application of the global spatial attention mechanism of~\cite{Wang02}.

\autoref{tab:rel} attempts to categorize related work on attention according to whether attention is local or global, spatial or channel, whether it is learned and whether it is applied to instance-level image retrieval. We observe that all methods limit to one or two forms of attention only. Of those studies that focus on image retrieval, many are not learned~\cite{JimenezAN17, gu2018attention, Kalantidis01}, and of those that are, some are designed for local descriptors~\cite{Noh01,tolias2020learning}.

By contrast, we provide a comprehensive study of \emph{all forms} of attention, global and local, spatial and channel, to obtain a learned representation in the form of a tensor that can be used in any way. We spatially pool it into a global descriptor and we study the relative gain of different forms of attention in image retrieval.

\section{Global-local attention}
\label{sec:method}

We design a \emph{global-local attention module} (GLAM), which is attached at the end of a backbone network. \autoref{fig:glam} illustrates its main components. We are given a $c\times h \times w$ feature tensor $\vF$, where $c$ is the number of channels, and $h \times w$ is the spatial resolution. Local attention collects context from the image and applies pooling to obtain a $c \times 1 \times 1$ \emph{local channel attention map} $\vA_c^l$ and a $1 \times h \times w$ \emph{local spatial attention map} $\vA_s^l$. Global attention allows interaction between channels, resulting in a $c \times c$ \emph{global channel attention map} $\vA_c^g$, and between spatial locations, resulting in a $hw \times hw$ \emph{global spatial attention map} $\vA_s^g$. The feature maps produced by the two attention streams are combined with the original one by a learned fusion mechanism into the \emph{global-local attention feature map} $\vF^{gl}$ before being spatially pooled into a global image descriptor.

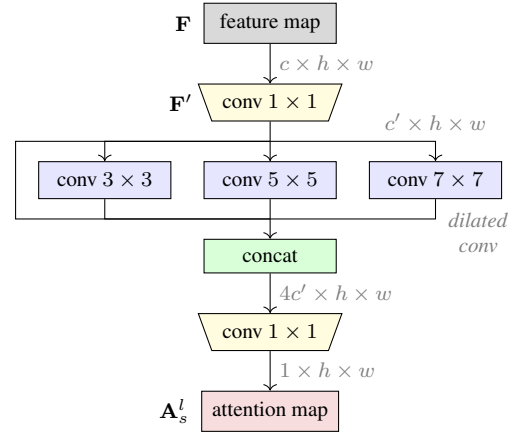
\begin{figure}
\centering
\begin{tikzpicture}[
	font={\footnotesize},
]
\matrix[
	row sep=15pt,column sep=10pt,cells={scale=1,},
]{
	\& \node[inp](i){feature map}; \\
	\& \node[down](d1){conv $1 \times 1$}; \\
	\node[fix](c3){conv $3 \times 3$}; \&
	\node[fix](c5){conv $5 \times 5$}; \&
	\node[fix](c7){conv $7 \times 7$}; \\
	\& \node[cat](ca){concat}; \\
	\& \node[down](d2){conv $1 \times 1$}; \\
	\& \node[outp](o){attention map}; \\
};
\coordinate(s) at($(d1.south)!.5!(c5.north)$); 
\coordinate(m) at($(c5.south)!.5!(ca.north)$); 
\node(left)[left=5pt of c3.west]{};
\coordinate(l) at(left.center);
\draw[->]
	(i) edge node[dim,midway,right]{$c \times h \times w$} (d1)
	(d1) edge (c5)
	(c5) edge (ca)
	(ca) edge node[dim,midway,right]{$4c' \times h \times w$} (d2)
	(d2) edge node[dim,midway,right]{$1 \times h \times w$} (o)
	;
\draw[->] (s) -| (c3);
\draw[->] (s) -| node[dim,midway,above]{$c' \times h \times w$} (c7);
\draw (s) -| (l) (l) |- (m);
\draw (c3) |- (m) (c7) |- (m);
\node[dim,below right=2pt of c7.south](dil) {\emph{dilated} \\ \emph{conv}};
\node[left=1pt of i] {$\vF$};
\node[left=1pt of d1] {$\vF'$};
\node[left=1pt of o] {$\vA_s^l$};
\end{tikzpicture}
\caption{Local spatial attention. Convolutional layers in blue implemented by dilated convolutions with kernel size $3 \times 3$ and dilation factors $1,3,5$.}
\label{fig:fig3}
\end{figure}

\subsection{Local attention}
\label{sec:local}

We extract an 1D channel and a 2D spatial attention map to weigh the feature map in the corresponding dimensions.

\paragraph{Local channel attention}

Following ECA-Net~\cite{wang01}, this attention captures local channel information. As shown in \autoref{fig:fig4}, we are given a $c\times h\times w$ feature tensor $\vF$ from our backbone. We first reduce it to a $c \times 1 \times 1$ tensor by \emph{global average pooling} (GAP). Channel attention is then captured by a 1D convolution of kernel size $k$ along the channel dimension, where $k$ controls the extent of cross-channel interaction. This is followed by a sigmoid function, resulting in the $c\times 1\times 1$ \emph{local channel attention map} $\vA_c^l$.


\paragraph{Local spatial attention}

Inspired by the inception module~\cite{Szegedy01} and similar to~\cite{Kim01}, this attention map captures local spatial information at different scales. As shown in \autoref{fig:fig3},
given the same $c\times h\times w$ feature tensor $\vF$ from our backbone, we obtain a new tensor $\vF'$ with channels reduced to $c'$, using a ${1 \times 1}$ convolution. We then extract local spatial contextual information using convolutional filters of kernel size ${3\times 3}$, ${5\times 5}$, and ${7\times 7}$, which are efficiently implemented by ${3\times 3}$ dilated convolutions~\cite{chen2017rethinking,Yu_2017_CVPR} with dilation parameter 1, 2, and 3 respectively. The resulting features, along with one obtained by ${1\times 1}$ convolution on $\vF'$, are concatenated into a $4c' \times h \times w$ tensor. Finally, we obtain the $1 \times h \times w$ \emph{local spatial attention map} $\vA_s^l$ by a ${1\times 1}$ convolution that reduces the channel dimension to $1$.

The middle column of \autoref{fig:fig7} shows heat maps of local spatial attention, localizing target objects in images.


\paragraph{Local attention feature map}

We use the local channel attention map $\vA_c^l$ to weigh $\vF$ in the channel dimension
\begin{equation}
	\vF_c^l \defn \vF \odot \vA_c^l + \vF.
\label{eq:eq3}
\end{equation}
We then use local spatial attention map $\vA_s^l$ to weigh $\vF_c^l$ in the spatial dimensions, resulting in the $c \times h \times w$ \emph{local attention feature map}
\begin{equation}
	\vF^l = \vF_c^l \odot \vA_s^l + \vF_c^l.
\label{eq:eq3-1}
\end{equation}
Here, $\vA \odot \vB$ denotes an element-wise multiplication of tensors $\vA$ and $\vB$, with broadcasting when one tensor is smaller. We adopt the choice of applying channel followed by spatial attention from \emph{convolutional block attention module} CBAM~\cite{woo01}. However, apart from computing $\vA_s^l$ at different scales, both attention maps are obtained from the original tensor $\vF$ rather than sequentially. In addition, both~\eq{eq3} and~\eq{eq3-1} include residual connections, while CBAM includes a single residual connection over both steps.

\begin{figure}
\centering
\begin{tikzpicture}[
	font={\footnotesize},
]
\matrix[
	row sep=15pt,column sep=10pt,cells={scale=1,},
]{
	\node[inp](i){feature map}; \&
	\node[gap](g){GAP}; \\
	\& \node[fix](c2){conv1d($k$)};
	\& \node[fix](c3){conv1d($k$)}; \\
	\& \node[fun](s1){$\sigmoid$};
	\& \node[fun](s2){$\sigmoid$}; \\
	\& \node[op](m1){$\times$}; \\
	\node[op](m2){$\times$}; \&
	\node[fun](sm){$\softmax$}; \\
	\node[outp](o){attention feature map}; \\
};
\coordinate(s) at($(g.south)!.5!(c2.north)$); 
\draw[->]
	(i) edge (g)
	(g)  edge (c2)
	(c2) edge node[dim,midway,right]{$1 \times c$} (s1)
	(c3) edge node[dim,midway,right]{$1 \times c$} (s2)
	(s1) edge node[dim,midway,right]{$1 \times c$} node[key,midway,left]{$\vQ_c$} (m1)
	(m1) edge node[dim,midway,right]{$c \times c$} (sm)
	(i) edge node[dim,midway,right]{$hw \times c$} node[key,midway,left]{$\vV_c$} (m2)
	(sm) edge node[midway,above]{$\vA_c^g$} (m2)
	(m2) edge node[dim,midway,right]{$c \times h \times w$} (o)
	;
\draw[->] (s) -| node[dim,midway,above]{$1 \times c$} (c3);
\draw[->] (s2) |- node[dim,near start,right]{$1 \times c$} node[key,near start,left]{$\vK_c$} (m1);
\node[left=1pt of i]{$\vF$};
\node[left=1pt of o]{$\vG_c$};
\end{tikzpicture}
\caption{Global channel attention.}
\label{fig:fig6}
\end{figure}
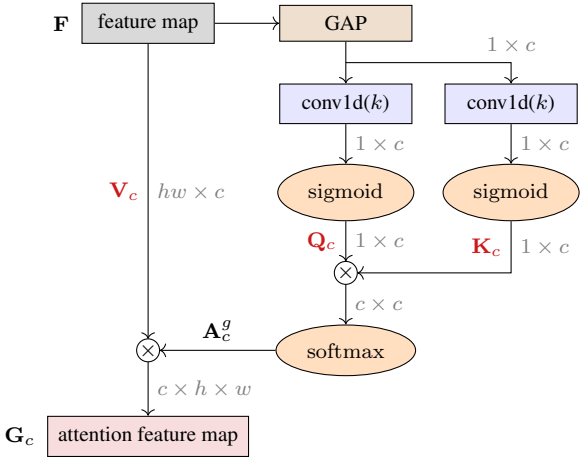


\subsection{Global attention}
\label{sec:global}

We extract two matrices capturing global pairwise channel and spatial interaction to weigh the feature map.


\paragraph{Global channel attention}

We introduce a \emph{global channel attention} mechanism that captures global channel interaction. This mechanism is based on the non-local neural network~\cite{Wang02}, but with the idea of 1D convolution from ECA-Net~\cite{wang01}. As shown in \autoref{fig:fig6}, we are given the $c\times h\times w$ feature tensor $\vF$ from our backbone. We apply GAP and squeeze spatial dimensions, followed by a 1D convolution of kernel size $k$ and a sigmoid function, to obtain $1 \times c$ \emph{query} $\vQ_c$ and \emph{key} $\vK_c$ tensors. The \emph{value} tensor $\vV_c$ is obtained by mere reshaping of $\vF$ to $hw \times c$, without GAP. Next, we form the outer product of $\vK_c$ and $\vQ_c$, followed by softmax over channels to obtain a $c \times c$ \emph{global channel attention map}
\begin{equation}
	\vA_c^g = \softmax({\vK_c}\tran \vQ_c).
\label{eq:eq6-1}
\end{equation}
Finally, this attention map is multiplied with $\vV_c$ and the matrix product $\vV_c \vA_c^g$ is reshaped back to $c \times h \times w$ to give the \emph{global channel attention feature map} $\vG_c$. In GSoP~\cite{Gao_2019_CVPR} and A$^2$-Net~\cite{ChenKLYF18}, a $c \times c$ global channel attention map is obtained by multiplication of $hw \times c$ matrices; \eq{eq6-1} is more efficient, using only an outer product of $1 \times c$ vectors.

\begin{figure}
\centering
\begin{tikzpicture}[
	font={\footnotesize},
]
\matrix[
	row sep=15pt,column sep=10pt,cells={scale=1,},
]{
	\& \node[inp](i){feature map}; \\
	\node[down](c1){conv $1 \times 1$}; \&
	\node[down](c2){conv $1 \times 1$}; \&
	\node[down](c3){conv $1 \times 1$}; \\
	\& \node[op](m1){$\times$}; \\
	\node[op](m2){$\times$}; \&
	\node[fun](sm){$\softmax$}; \\
	\node[up](up){conv $1 \times 1$}; \\
	\node[outp](o){attention feature map}; \\
};
\coordinate(s) at($(i.south)!.5!(c2.north)$); 
\draw[->]
	(i) edge (c2)
	(c2) edge node[dim,midway,right]{$c' \times hw$} node[key,midway,left]{$\vQ_s$} (m1)
	(m1) edge node[dim,midway,right]{$hw \times hw$} (sm)
	(up) edge node[dim,midway,right]{$c \times h \times w$} (o)
	(c1) edge node[dim,midway,right]{$c' \times hw$} node[key,midway,left]{$\vV_s$} (m2)
	(m2) edge node[dim,midway,right]{$c' \times h \times w$} (up)
	(sm) edge node[midway,above]{$\vA_s^g$} (m2)
	;
\draw[->] (s) -| (c1);
\draw[->] (s) -| node[dim,midway,above]{$c \times h \times w$} (c3);
\draw[->] (c3) |- node[dim,near start,right]{$c' \times hw$} node[key,near start,left]{$\vK_c$} (m1);
\node[left=1pt of i]{$\vF$};
\node[left=1pt of o]{$\vG_s$};
\end{tikzpicture}
\caption{Global spatial attention.}
\label{fig:fig5}
\end{figure}
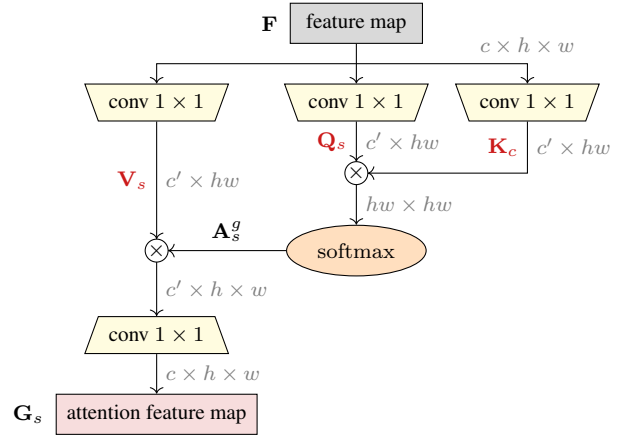

\paragraph{Global spatial attention}

Since ordinary convolution applies only a local neighborhood at a time, it cannot capture global contextual information. Thus, we apply \emph{non-local filtering}~\cite{Wang02}, which is a form of \emph{self-attention}~\cite{Vaswani01} in the spatial dimensions. As shown in \autoref{fig:fig5}, we are given the same $c\times h\times w$ feature tensor $\vF$ from our backbone. By using three $1\times 1$ convolutions, which reduce channels to $c'$, and flattening spatial dimensions to $hw$, we obtain $c' \times hw$ \emph{query} $\vQ_s$, \emph{key} $\vK_s$, and \emph{value} $\vV_s$ tensors, where each column is a feature vector corresponding to a particular spatial location. We capture pairwise similarities of these vectors by matrix multiplication of $\vK_s$ and $\vQ_s$, followed by softmax over locations to obtain a $hw \times hw$ \emph{global spatial attention map}:
\begin{equation}
	\vA_s^g = \softmax(\vK_s\tran \vQ_s).
\label{eq:eq4-1}
\end{equation}
This attention map is multiplied with $\vV_s$ and the matrix product $\vV_s \vA_s^g$ is reshaped back to $c' \times h \times w$ by expanding the spatial dimensions. Finally, using a ${1\times 1}$ convolution, which increases channels back to $c$, we obtain the $c \times h\times w$ \emph{global spatial attention feature map} $\vG_s$.

\begin{figure}
\centering
\small
\setlength{\tabcolsep}{2pt}
\newcommand{\heat}[1]{%
	\fig[.28]{heatmap/#1/src.png} &
	\fig[.28]{heatmap/#1/l.png} &
	\fig[.28]{heatmap/#1/g.png} \\
}
\begin{tabular}{ccc}
	\heat{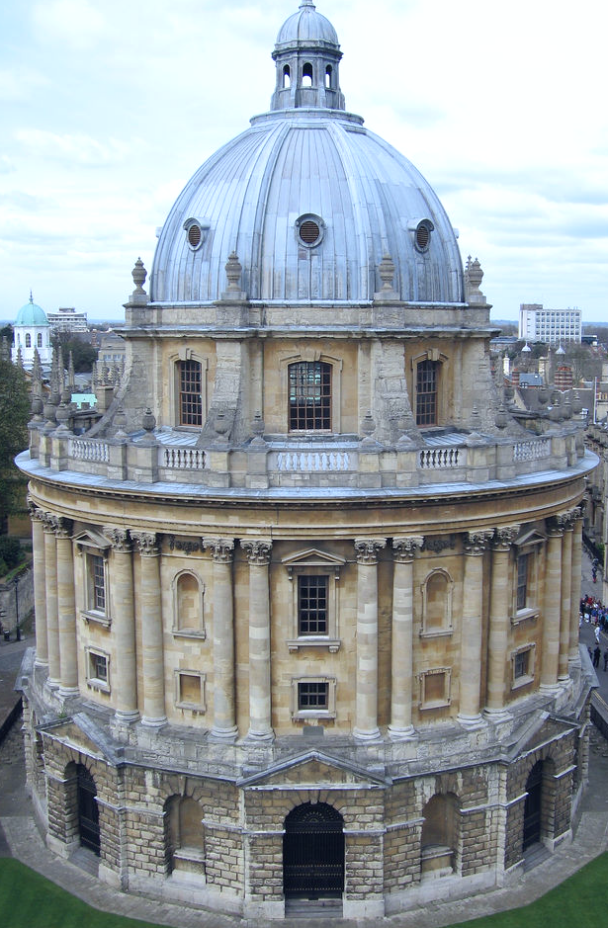}
	\heat{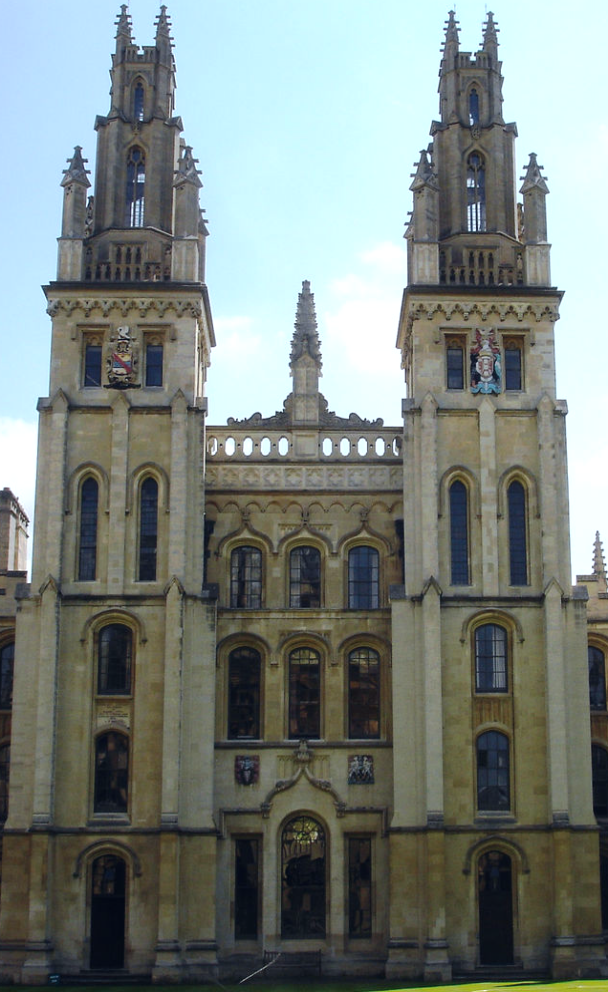}
	\heat{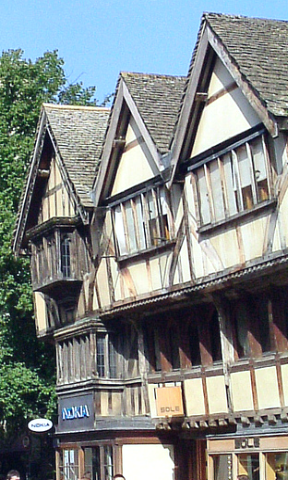}
	(a) input &
	(b) local &
	(c) global
\end{tabular}
\caption{\emph{Local and global spatial attention}. Left: input images. Middle: local spatial attention heat maps. Right: global spatial attention heat maps. Red (blue) means higher (lower) attention weight.}
\label{fig:fig7}
\end{figure}

The right column of \autoref{fig:fig7} shows heat maps for global spatial attention, localizing target objects in images.


\paragraph{Global attention feature map}

We use the global channel attention feature map $\vF_c$ to weigh $\vF$ element-wise
\begin{equation}
	\vF_c^g = \vF \odot \vG_c.
\label{eq:eq8}
\end{equation}
We then use global spatial attention feature map $\vG_s$ to weigh $\vF_c^g$ element-wise, resulting in the $c \times h \times w$ \emph{global attention feature map}
\begin{equation}
	\vF^g = \vF_c^g \odot \vG_s + \vF_c^g.
\label{eq:eq8-1}
\end{equation}
Similarly to $\mathbf{F}^{l}$ in~\eq{eq3} and~\eq{eq3-1}, we apply channel attention first, followed by spatial attention. However, unlike \eq{eq3}, there is no residual connection in~\eq{eq8}. This choice is supported by early experiments.

\subsection{Global-local attention}
\label{sec:embed}

\paragraph{Feature fusion}

As shown in \autoref{fig:glam}, we combine the local and global attention feature maps, $\vF^l$ and $\vF^g$, with the original feature $\vF$. While concatenation and summation are common operations for feature combination, we use a weighted average with weights $w_l$, $w_g$, $w$ respectively, obtained by softmax over three learnable scalar parameters, to obtain a $c \times h \times w$ \emph{global-local attention feature map}
\begin{equation}
	\vF^{gl} = w_l \vF^l + w_g \vF^l + w \vF.
\label{eq:eq10}
\end{equation}
EfficientDet~\cite{Tan01} has shown that this is the most effective, among a number of choices, for fusion of features across different scales.


\paragraph{Pooling}

We apply GeM~\cite{Radenovic01}, a learnable spatial pooling mechanism, to feature map $\vF^{gl}$~\eq{eq10}, followed by a fully-connected (FC) layer with dropout and batch normalization. The final embedding is obtained by $\ell_2$-normalization.

\section{Experiments}
\label{sec:exp}

\begin{figure*}
\centering
\fig{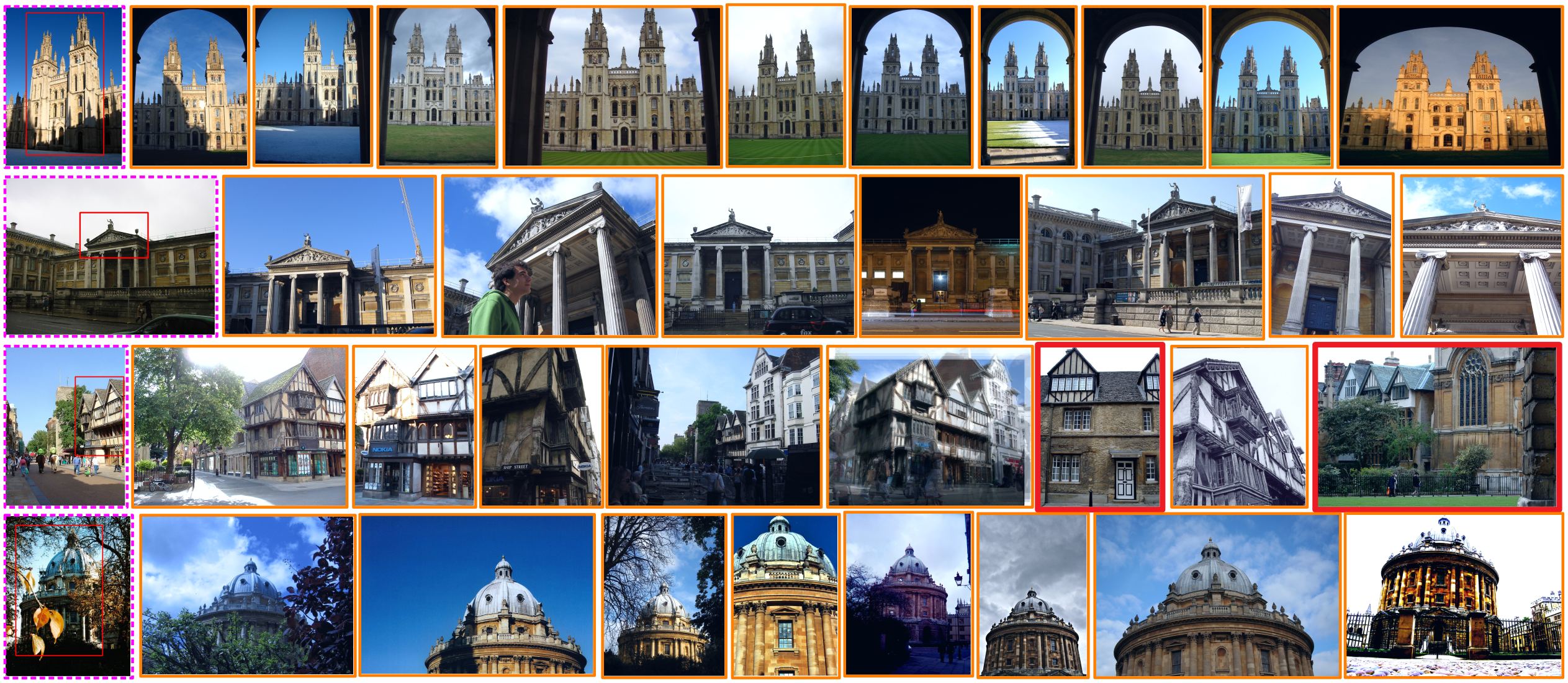}
\caption{Examples of our ranking results. In each row, the first image on the left (pink dotted outline) is a query image with a target object (red crop box), and the following are the top ranking images for the query. Orange solid outline: positive images for the query; red solid outline: negative.}
\label{fig:fig8}
\end{figure*}

\subsection{Datasets}

\paragraph{Training set}

There are a number of open landmark datasets commonly used for training in image retrieval studies, including \emph{neural code} (NC)~\cite{Babenko01}, \emph{neural code clean} (NC-clean)~\cite{Gordo01}, as well as Google Landmarks v1 (GLDv1)~\cite{Noh01} and v2 (GLDv2)~\cite{Weyand01}. \autoref{tab:table1} shows relevant statistics. These datasets can be categorized into noisy and clean. The clean sets were obtained from the original noisy sets for more effective training~\cite{Gordo01, Weyand01}. The original noisy datasets are much larger, but they have high intra-class variability. Each class can include visually dissimilar images such as exterior and interior views of a building or landmark, including floor plans and paintings inside. The clean datasets focus on views directly relevant to landmark recognition but have a much smaller number of images.


\paragraph{Evaluation set and metrics}

We use four common evaluation datasets for landmark image retrieval: Oxford5k (\oxf5k)~\cite{Philbin01}, Paris6k (\paris6k)~\cite{Philbin02}, as well as Revisited Oxford (\roxf~or \rox) and Paris (\rpar~or \rpa)~\cite{RITAC18}. \roxf~and \rpar~are used with and without one million distractors (\r1m)~\cite{Ng01} and evaluated using the Medium and Hard protocols~\cite{RITAC18}. We evaluate using \emph{mean Average Precision} (mAP) and \emph{mean precision at} 10 (mP@10).


\subsection{Implementation details}

We train on 8 TITAN RTX 2080Ti GPUs. All models are pre-trained on ImageNet~\cite{Russakovsky01} and implemented in PyTorch \cite{Paszke01}. For fair comparisons, we set a training environment similar to the those  of compared studies~\cite{Yokoo01, Weyand01, Ng01, RITAC18}. We employ ResNet101~\cite{Zhang01} as a backbone model. The kernel size $k$ of ECANet in \autoref{sec:local} is set to 3. The parameter $p$ of GeM in \autoref{sec:embed} is set to 3 and the dimension $d$ of final embeddings to 512. We adopt ArcFace~\cite{Deng01}, a cosine-softmax based loss, with a margin of 0.3. We use stochastic gradient descent with initial learning rate $10^{-3}$, momentum 0.9 and weight decay $10^{-5}$.

We adopt the batch sampling of Yokoo \etal~\cite{Yokoo01} where mini-batch samples with similar aspect ratios are resized to a particular size. Here, we use a batch size of 64. For image augmentation, we apply scaling, random cropping, and varied illumination. At inference, we apply a multi-resolution representation~\cite{Gordo01} to query and database images.

Our method is denoted as GLAM (\emph{global-local attention module}). Using the backbone model alone is referred to as \emph{baseline}. It is compatible with recent models based on ResNet101-GeM trained with ArcFace~\cite{Weyand01, Ng01}. Adding our local attention (\autoref{sec:local}) to the baseline model is denoted \emph{+local}, while adding our global attention (\autoref{sec:global}) is denoted \emph{+global}. Since we focus on representation learning, we do not consider post-processing methods like geometry-based re-ranking \cite{Noh01, simeoni2019local, Weyand01} or graph-based re-ranking~\cite{Donoser01, iscen2017efficient, Yang01}.

\begin{table}
\centering
\small
\begin{tabular}{lcc} \toprule
\Th{Train Set} & \Th{\#Images} & \Th{\#Classes} \\ \midrule
NC-noisy &  213,678 & 672 \\
NC-clean & 27,965 & 581 \\
SfM-120k & 117,369 & 713 \\
GLDv1-noisy & 1,225,029  &  14, 951 \\
GLDv2-noisy & 4,132,914  &  203,094 \\
GLDv2-clean & 1,580,470 & 81,313 \\
\bottomrule
\end{tabular}
\caption{Statistics of different training sets.}
\label{tab:table1}
\end{table}

\subsection{Benchmarking}
\label{sec:SOTA}

\begin{table}
\centering
\scriptsize
\setlength{\tabcolsep}{0.6pt}
\begin{tabular}{l*{8}{c}} \toprule
\mr{2}{\Th{Method}} & \mr{2}{\Th{Train Set}} & \mr{2}{\Th{dim}} & \mr{2}{\Th{Oxf5k}} & \mr{2}{\Th{Par6k}} & \mc{2}{\Th{$\cR$Medium}} & \mc{2}{\Th{$\cR$Hard}} \\ \cmidrule(l){6-9}
 & & & & & \rox & \rpa & \rox & \rpa \\ \midrule
GeM-Siamese \cite{Radenovic01, RITAC18} & SfM-120k    & 2048 & 87.8 & 92.7 & 64.7 & 77.2 & 38.5 & 56.3 \\
SOLAR~\cite{Ng01}                       & GLDv1-noisy & 2048 & --   & --   & 69.9 & 81.6 & 47.9 & 64.5 \\
GLDv2~\cite{Weyand01}                   & GLDv2-clean & 2048 & --   & --   & 74.2 & 84.9 & 51.6 & 70.3 \\
\midrule
GLAM (Ours) &  NC-clean & 512 &  77.8 & 85.8 & 51.6 &  68.1 &  20.9 & 44.7 \\
 &  GLDv1-noisy & 512 & 92.8 & 95.0 & \ok{\tb{73.7}}  &  \ok{\tb{83.5}}  &  \ok{\tb{49.8}} & \ok{\tb{69.4}} \\
 &  GLDv2-noisy & 512 & 93.3 & 95.3 & 75.7 & 86.0 & 53.1 & 73.8 \\
 &  GLDv2-clean & 512 & \red{\tb{94.2}} & \red{\tb{95.6}} & \red{\tb{78.6}} & \red{\tb{88.5}} & \red{\tb{60.2}} & \red{\tb{76.8}} \\ \bottomrule
\end{tabular}
\caption{mAP comparison of our best model (baseline+local+global) trained on different \emph{training sets} against \cite{Weyand01,Ng01}. All models use ResNet101-GeM. Red: best results. Blue: GLAM higher than SOLAR~\cite{Ng01} on GLDv1-noisy.}
\label{tab:table11}
\end{table}

\begin{table*}
\centering
\scriptsize
\setlength{\tabcolsep}{2pt}
\begin{tabular}{l|cc|cc|cccccccc|cccccccc} \toprule
	\mr{3}{\Th{Method}} & \mr{3}{\Th{Train Set}} & \mr{3}{\Th{Dim}} & \mca{2}{c|}{\Th{Base}} & \mca{8}{c|}{\Th{Medium}} & \mc{8}{\Th{Hard}} \\
	                                                   &             &          & \oxf5k & \paris6k & \mc{2}{\rox} & \mc{2}{+\r1m} & \mc{2}{\rpa} & \multicolumn{2}{c|}{+\r1m} & \mc{2}{\rox} & \mc{2}{+\r1m} & \mc{2}{\rpa} & \mc{2}{+\r1m} \\
	                                                   &             &          & mAP &  mAP & mAP & mP & mAP & mP & mAP & mP & mAP & mP & mAP & mP & mAP & mP & mAP & mP & mAP & mP \\ \midrule
	SPoC-V16 \cite{Babenko03, RITAC18}                 & [O]         & 512      & 53.1$^*$ & -- & 38.0 & 54.6 & 17.1 & 33.3 & 59.8 & 93.0 & 30.3 & 83.0 & 11.4 & 20.9 & 0.9 & 2.9 & 32.4 & 69.7 & 7.6 & 30.6 \\
	SPoC-R101 \cite{RITAC18}                           & [O]         & 2048     & -- & -- & 39.8 & 61.0 & 21.5 & 40.4 & 69.2 & 96.7 & 41.6 & 92.0 & 12.4 & 23.8 & 2.8 & 5.6 & 44.7 & 78.0 & 15.3 & 54.4 \\
	CroW-V16 \cite{Kalantidis01,RITAC18}               & [O]         & 512      & 70.8 & 79.7 & 41.4 & 58.8 & 22.5 & 40.5 & 62.9 & 94.4 & 34.1 & 87.1 & 13.9 & 25.7 & 3.0 & 6.6 & 36.9 & 77.9 & 10.3 & 45.1 \\
	CroW-R101 \cite{RITAC18}                           & [O]         & 2048     & -- & -- & 42.4 & 61.9 & 21.2 & 39.4 & 70.4 & 97.1 & 42.7 & 92.9 & 13.3 & 27.7 & 3.3 & 9.3 & 47.2 & 83.6 & 16.3 & 61.6 \\
	MAC-V16-Siamese \cite{Radenovi01, RITAC18}         & [O]         & 512      & 80.0 & 82.9 & 37.8 & 57.8  & 21.8  & 39.7 & 59.2 & 93.3  & 33.6 & 87.1 & 14.6  & 27.0  & 7.4  & 11.9 & 35.9  & 78.4  & 13.2 & 54.7 \\
	MAC-R101-Siamese \cite{RITAC18}                    & [O]         & 2048     & -- & -- & 41.7 & 65.0 & 24.2 & 43.7 & 66.2 & 96.4 & 40.8 & 93.0 & 18.0 & 32.9 & 5.7 & 14.4 & 44.1 & 86.3 & 18.2 & 67.7 \\
	RMAC-V16-Siamese \cite{Radenovi01, RITAC18}        & [O]         & 512      & 80.1 & 85.0 & 42.5 & 62.8 & 21.7 & 40.3 & 66.2 & 95.4 & 39.9 & 88.9 & 12.0 & 26.1 & 1.7 & 5.8 & 40.9 & 77.1 & 14.8 & 54.0 \\
	RMAC-R101-Siamese \cite{RITAC18}                   & [O]         & 2048     & -- & -- & 49.8 & 68.9 &  29.2 &  48.9 &  74.0 &  97.7 &  49.3 &  93.7 &  18.5 & 32.2 &  4.5 &  13.0 &  52.1 &  87.1 &  21.3 &  67.4 \\
	RMAC-R101-Triplet \cite{Gordo01, RITAC18}          & NC-clean    & 2048     & 86.1 & \tb{94.5} & 60.9 & 78.1 & 39.3 & 62.1 & 78.9 & 96.9 & 54.8 & 93.9 & 32.4 & 50.0 & 12.5 & 24.9 & 59.4 & 86.1 & 28.0 & 70.0 \\
	GeM-R101-Siamese \cite{Radenovic01, RITAC18}       & SfM-120k    & 2048     & \tb{87.8} & 92.7 & 64.7 & 84.7 & 45.2 & 71.7 & 77.2 & \red{\tb{98.1}} & 52.3 & \red{\tb{95.3}} & 38.5 & 53.0 & 19.9 & 34.9 & 56.3 & 89.1 & 24.7 & 73.3 \\
	AGeM-R101-Siamese \cite{gu2018attention}           & SfM-120k    & 2048     & -- & -- & 67.0 & -- & -- & -- & 78.1 & --  & -- & -- & 40.7 & -- & -- & -- & 57.3 & -- & -- & -- \\
	SOLAR-GeM-R101-Triplet/SOS \cite{Ng01}             & GLDv1-noisy & 2048     & -- & -- & 69.9 & \tb{86.7} & 53.5 & \tb{76.7} & 81.6  & 97.1 & 59.2 & 94.9 & 47.9 & \tb{63.0} & 29.9 & \tb{48.9} &  64.5 & \tb{93.0} & 33.4 & \tb{81.6} \\
	DELG-GeM-R101-ArcFace \cite{ECCV2020_912}          & GLDv1-noisy & 2048     & -- & -- & 73.2 & -- & \tb{54.8} & -- & 82.4 & --  & \tb{61.8} & -- & 51.2 & -- & \tb{30.3} & -- & 64.7 & -- & \tb{35.5} & -- \\
	GeM-R101-ArcFace \cite{Weyand01}                   & GLDv2-clean & 2048     & -- & -- & \tb{74.2} & -- & -- & -- & \tb{84.9} & --  & -- & -- &  \tb{51.6}  & -- & -- & -- &  \tb{70.3} & -- & -- & -- \\
	\midrule
	GLAM-GeM-R101-ArcFace baseline   & GLDv2-clean & 512 & \ok{\tb{91.9}} & \ok{\tb{94.5}} & 72.8 & \ok{\tb{86.7}} &  \ok{\tb{58.1}} & \ok{\tb{78.2}} & 84.2 & 95.9 &  \ok{\tb{63.9}} &  93.3 &  49.9 & 62.1 & \ok{\tb{31.6}} & \ok{\tb{49.7}} & 69.7 & 88.4 & \ok{\tb{37.7}} & 73.7  \\
	+local                           & GLDv2-clean & 512 & \ok{\tb{91.2}} & \ok{\tb{95.4}} & 73.7 & \ok{\tb{86.2}} & \ok{\tb{60.5}} & \ok{\tb{77.4}} & \ok{\tb{86.5}} & 95.6 & \ok{\tb{68.0}} & 93.9  &  \ok{\tb{52.6}} & \ok{\tb{65.3}} & \ok{\tb{36.1}} & \ok{\tb{55.6}} & \ok{\tb{73.7}} & 89.3 & \ok{\tb{44.7}} & 79.1  \\
	+global                          & GLDv2-clean & 512 & \ok{\tb{92.3}} & \ok{\tb{95.3}} & \ok{\tb{77.2}} & \ok{\tb{87.0}} & \ok{\tb{63.8}} & \ok{\tb{79.3}} & \ok{\tb{86.7}} & 95.4 & \ok{\tb{67.8}} & 93.7  & \ok{\tb{57.4}} & \ok{\tb{69.6}} & \ok{\tb{38.7}} & \ok{\tb{57.9}} & \ok{\tb{75.0}} & 89.4 & \ok{\tb{45.0}} & 77.0  \\
	+global+local                    & GLDv2-clean & 512 & \red{\tb{94.2}} & \red{\tb{95.6}} & \red{\tb{78.6}} & \red{\tb{88.2}} & \red{\tb{68.0}} & \red{\tb{82.4}} & \red{\tb{88.5}}  & 97.0 &  \red{\tb{73.5}} & 94.9 & \red{\tb{60.2}} & \red{\tb{72.9}} & \red{\tb{43.5}} & \red{\tb{62.1}} & \red{\tb{76.8}} & \red{\tb{93.4}} & \red{\tb{53.1}} & \red{\tb{84.0}} \\
	\bottomrule
\end{tabular}
\caption{mAP comparison of our GLAM against SOTA methods based on global descriptors without re-ranking. V16: VGG16; R101: ResNet101. [O]: Off-the-shelf (pre-trained on ImageNet). $^*$: dimension $d=256$~\cite{Babenko03}. mP: mP@10. Red: best results. Black bold: best previous methods. Blue: GLAM higher than previous methods. Weyand \etal~\cite{Weyand01} is the only model other than ours trained on GLDv2-clean, while~\cite{Ng01} is trained on GLDv1-noisy and compared in \autoref{tab:table11}.}
\label{tab:table_exp_all}
\end{table*}

\paragraph{Noisy \vs clean training sets}

We begin by training our best model (baseline+local+global)
on all training sets of \autoref{tab:table1}, except NC-noisy because some images are currently unavailable. As shown in \autoref{tab:table11}, even though GLDv2-noisy has 2.6 times more images than GLDv2-clean, the latter is superior by a large margin. This shows that, in training, a cleaner dataset can be more important than a larger one. By contrast, NC-clean has the worst performance despite being clean, aparently because it is too small. To achieve best possible performance, we use GLDv2-clean as a training set in the remaining experiments.


\paragraph{Comparisons on same training set}

It is common to compare methods regardless of training sets as more become available, \eg,~\cite{RITAC18, Ng01}. Since GLDv2-clean is relatively new, Weyand \etal~\cite{Weyand01}, which introduced the dataset, is the only study that has trained the same backbone with the same settings (ResNet101-GeM with ArcFace) on GLDv2-clean. Our baseline is lower than~\cite{Weyand01}, because our dimensinality is 512, while other models based on ResNet101 use 2048. Yet, \autoref{tab:table11} shows that our best model trained on GLDv2-clean outperforms~\cite{Weyand01} by a large margin. But the most important comparison is with SOLAR~\cite{Ng01}, also based on self-attention, which has trained ResNet101-GeM on GLDv1-noisy. On this training set, our best model clearly outperforms~\cite{Ng01} despite lower dimensionality.


\paragraph{Comparison with state of the art}

\autoref{tab:table_exp_all} shows the performance of four variants of our model, \ie baseline with or without local/global attention, and compares them against state-of-the-art (SOTA) methods based on global descriptors without re-ranking on the complete set of benchmarks, including distractors. Both local and global attention bring significant gain over the baseline. The effect of global is stronger, while the gain of the two is additive in the combination. The best results are achieved by the global-local attention network (baseline+global+local). With this model, we outperform previous best methods on most benchmarks except mP@10 on \rpar~(medium) and \rpar$+$\r1m~(medium), where we are outperformed by~\cite{Radenovic01, RITAC18}. These results demonstrate that our approach is effective for landmark image retrieval. \autoref{fig:fig8} shows some examples of our ranking results.

\subsection{Ablation study}
\label{sec:Ablation}

\begin{table}
\centering
\small
\setlength{\tabcolsep}{1.5pt}
\begin{tabular}{l*{6}{c}} \toprule
\mr{2}{\Th{Method}} & \mr{2}{\Th{Oxf5k}} & \mr{2}{\Th{Par6k}} & \mc{2}{\Th{$\cR$Medium}} & \mc{2}{\Th{$\cR$Hard}} \\ \cmidrule(l){4-7}
 & & & \rox & \rpa & \rox & \rpa \\ \midrule
GLAM baseline  & 91.9 & 94.5 & 72.8 &  84.2 & 49.9 & 69.7 \\
+local-channel & 91.3 & 95.3 & 72.2 & 85.8 & 48.3 & 73.1 \\
+local-spatial & 91.0 & 95.1 & 72.1 & 85.3 & 48.3 &  71.9 \\
+local & 91.2 & 95.4 & 73.7 & 86.5 & 52.6 & 75.0 \\
+global-channel & 92.5 & 94.4 & 73.3 & 84.4 & 49.8 & 70.1 \\
+global-spatial & 92.4 & 95.1 & 73.2 & 86.3 & 50.0 & 72.7 \\
+global & 92.3 & 95.3 & 77.2 & 86.7 & 57.4 & 75.0 \\
+global+local  & \tb{94.2} & \tb{95.6} & \tb{78.6}  & \tb{88.5}  & \tb{60.2} & \tb{76.8} \\ \bottomrule
\end{tabular}
\caption{mAP comparison of spatial and channel variants of our local (+local, \autoref{sec:local}) and global (+global, \autoref{sec:local}) attention modules to the baseline.}
\label{tab:table10}
\end{table}

\begin{table}
\centering
\small
\setlength{\tabcolsep}{3.4pt}
\begin{tabular}{l*{6}{c}} \toprule
\mr{2}{\Th{Method}} & \mr{2}{\Th{Oxf5k}} & \mr{2}{\Th{Par6k}} & \mc{2}{\Th{$\cR$Medium}} & \mc{2}{\Th{$\cR$Hard}} \\ \cmidrule(l){4-7}
 & & & \rox & \rpa & \rox & \rpa \\ \midrule
CBAM style  & 93.8 & \tb{95.7} & 75.6 &  88.4 & 53.3 & \tb{76.8} \\
GLAM (Ours)  & \tb{94.2} & 95.6 & \tb{78.6}  & \tb{88.5}  & \tb{60.2} & \tb{76.8} \\ \bottomrule
\end{tabular}
\caption{mAP comparison between CBAM style and our local spatial attention.}
\label{tab:table7}
\end{table}

\begin{table}
\centering
\small
\setlength{\tabcolsep}{4.2pt}
\begin{tabular}{l*{6}{c}} \toprule
\mr{2}{\Th{Method}} & \mr{2}{\Th{Oxf5k}} & \mr{2}{\Th{Par6k}} & \mc{2}{\Th{$\cR$Medium}} & \mc{2}{\Th{$\cR$Hard}} \\ \cmidrule(l){4-7}
 & & & \rox & \rpa & \rox & \rpa \\ \midrule
Concatenate  & 89.5 & 95.1 & 73.6 & 86.5 & 54.0 & 73.7 \\
Sum (Ours) & \tb{94.2} & \tb{95.6} & \tb{78.6}  & \tb{88.5}  & \tb{60.2} & \tb{76.8} \\ \bottomrule
\end{tabular}
\caption{mAP comparison between weighted concatenation and weighted average for feature fusion.}
\label{tab:table5}
\end{table}

\begin{table}
\centering
\small
\setlength{\tabcolsep}{2.5pt}
\begin{tabular}{l*{6}{c}} \toprule
\mr{2}{\Th{Method}} & \mr{2}{\Th{Oxf5k}} & \mr{2}{\Th{Par6k}} & \mc{2}{\Th{$\cR$Medium}} & \mc{2}{\Th{$\cR$Hard}} \\ \cmidrule(l){4-7}
 & & & \rox & \rpa & \rox & \rpa \\ \midrule
Fixed-size  & 76.1 & 82.6 & 55.7 & 68.4 & 29.2 & 47.5 \\
Group-size (Ours) & \tb{94.2} & \tb{95.6} & \tb{78.6}  & \tb{88.5}  & \tb{60.2} & \tb{76.8} \\ \bottomrule
\end{tabular}
\caption{mAP comparison between fixed-size ($224 \times 224$) and group-size sampling methods.}
\label{tab:table4}
\end{table}

\begin{table}
\centering
\small
\setlength{\tabcolsep}{1.9pt}
\begin{tabular}{l*{7}{c}} \toprule
\mr{2}{\Th{Query}} & \mr{2}{\Th{Database}} & \mr{2}{\Th{Oxf5k}} & \mr{2}{\Th{Par6k}} & \mc{2}{\Th{$\cR$Medium}} & \mc{2}{\Th{$\cR$Hard}} \\ \cmidrule(l){5-8}
 & & & & \rox & \rpa & \rox & \rpa \\ \midrule
Single & Single & 93.3 & 95.2 & 76.9 & 87.1 & 58.6 & 74.7 \\
Multi & Single & 93.9 & 95.4 & 78.0 & 87.7 & 59.0 & 75.5 \\
Single & Multi & 93.6 & \tb{95.6} & 77.0 & 87.8 & 57.1 & 76.0 \\
Multi & Multi & \tb{94.2} & \tb{95.6} & \tb{78.6}  & \tb{88.5}  & \tb{60.2} & \tb{76.8} \\ \bottomrule
\end{tabular}
\caption{mAP comparison of using multiresolution representation (Multi) or not (Single) on query or database.}
\label{tab:table6}
\end{table}

Our ablation study uses the Google Landmark v2 clean dataset (GLDv2-clean)~\cite{Weyand01} for training, which is shown to be the most effective in \autoref{tab:table11}.


\paragraph{Effect of attention modules}

We ablate the effect of our local and global attention networks as well as their combination. \autoref{tab:table10} shows the results, which are more fine-grained than those of \autoref{tab:table_exp_all}. In particular, it shows the effect of the channel and spatial variants of both local and global attention. We observe that, when used alone, the channel and spatial variants of local attention are harmful in most cases. Even the combination, baseline+local, is not always effective. By contrast, when used alone, the channel and spatial variants of global attention are mostly beneficial, especially the latter. Their combination, baseline+global, is impressive, bringing gain of up to 7.5\%. Importantly, the combination baseline+global+local improves further by up to another 2.8\%. This result shows the necessity of local attention in the final model.


\paragraph{CBAM \vs our local spatial attention}

We experiment with the local spatial attention of CBAM~\cite{woo01}. CBAM applies average and max-pooling to input features and concatenates the two for spatial attention. We apply this variant to our local spatial attention module for comparison. For the CBAM style module, we keep the overall design of our module as shown in \autoref{fig:fig3}, but apply average and max-pooling to each of the four convolutional layer outputs before concatenation. \autoref{tab:table7} shows that the CBAM style module is considerably worse than ours on all benchmarks except Paris6k, where it is only slightly better.


\paragraph{Concatenation \vs sum for feature fusion}

We use a softmax-based weighted average of local and global attention feature maps with the original feature map~\eq{eq10}. Here, we compare this weighted average with weighted concatenation, where concatenation replaces the sum operation in~\eq{eq10}. As shown in \autoref{tab:table5}, the weighted average outperforms the weighted concatenation.


\paragraph{Fixed-size \vs group-size sampling}

Numerous studies have proposed methods for constructing batches according to image size for efficient training. For instance, Gordo \etal~\cite{Gordo01}, DELF~\cite{Noh01}, and Yokoo~\etal\cite{Yokoo01} employed different image sizes per batch for training instead of a single fixed size. We adopt the method of Yokoo \etal, which constructs a batch with images of similar aspect ratio, so that the images can be resized to a size with an aspect ratio that is similar to their own. We call this method \emph{group-size sampling}. \autoref{tab:table4} compares fixed-size ($224 \times 224$) with group-size sampling. We observe that maintaining aspect ratios by using dynamic input sizes is much more effective.


\paragraph{Multi-resolution}

We use the multi-resolution representation~\cite{Gordo01} for the final feature of an image at inference time. This method: (1) resizes an image into multiple scales; (2) extracts features from the resized images; and (3) averages the features to obtain the final feature of the image. The method is applied to both query and database images to enhance ranking results, especially for small target objects. \autoref{tab:table6} compares the four cases of applying this method or not to query or database images.

\section{Conclusion}
\label{sec:conclusion}

We have introduced a novel approach that extracts global and local contextual information using attention mechanisms for instance-level image retrieval. It is manifested as a network architecture consisting of global and local attention components, each operating on both spatial and channel dimensions. This constitutes a comprehensive study and empirical evaluation of all four forms of attention that have previously been studied only in isolation. Our findings indicate that the gain (or loss) brought by one form of attention alone strongly depends on the presence of the others, with the maximum gain appearing when all forms are present. The output is a modified feature tensor that can be used in any way, for instance with local feature detection instead of spatial pooling for image retrieval.

With the advent of \emph{vision transformers}~\cite{dosovitskiy2020image,2101.11986} and their recent application to image retrieval~\cite{2102.05644}, attention is expected to play a more and more significant role in vision. According to our classification, transformers perform global spatial attention alone. It is of great interest to investigate the role of the other forms of attention, where our approach may yield a basic building block of such architectures. One may even envision an extension to language models, where transformers originate from~\cite{VSP+17}.

{\small
\bibliographystyle{ieee_fullname}
\bibliography{main}
}

\end{document}